\documentclass{article}

\usepackage{arxiv}

\usepackage[utf8]{inputenc} % allow utf-8 input
\usepackage[T1]{fontenc}    % use 8-bit T1 fonts
\usepackage{hyperref}       % hyperlinks
\usepackage{url}            % simple URL typesetting
\usepackage{booktabs}       % professional-quality tables
\usepackage{amsfonts}       % blackboard math symbols
\usepackage{nicefrac}       % compact symbols for 1/2, etc.
\usepackage{microtype}      % microtypography
\usepackage{cleveref}       % smart cross-referencing
\usepackage{graphicx}
\usepackage{natbib}
\usepackage{doi}
\usepackage{float}
\usepackage{pgfplots}
\usepackage{comment}
\usepackage{todonotes}
\usepackage{subcaption}

\pgfplotsset{compat=1.18}

\title{Python Agent in Ludii}

\newif\ifuniqueAffiliation
\uniqueAffiliationtrue

\ifuniqueAffiliation % Standard variant of author block
\author{ %\href{https://orcid.org/0000-0000-0000-0000
{%\includegraphics[scale=0.06]{orcid.pdf}\hspace{1mm}
Izaias Saturnino de Lima Neto} \\
	Instituto de Informática\\
	Universidade Federal do Rio Grande do Sul\\
	Porto Alegre, RS 9500 \\
	\texttt{islneto@inf.ufrgs.br} \\
    \And
    {%\includegraphics[scale=0.06]{orcid.pdf}\hspace{1mm}
    Marco Antônio Athayde De Aguiar Vieira
    } \\
	Instituto de Informática\\
	Universidade Federal do Rio Grande do Sul\\
	Porto Alegre, RS 9500 \\
	\texttt{marco.vieira@inf.ufrgs.br} \\
    \And
    {%\includegraphics[scale=0.06]{orcid.pdf}\hspace{1mm}
    Anderson Rocha Tavares
    } \\
	Instituto de Informática\\
	Universidade Federal do Rio Grande do Sul\\
	Porto Alegre, RS 9500 \\
	\texttt{artavares@inf.ufrgs.br} \\
}
\else
\fi

\renewcommand{\undertitle}{Technical Report}
\hypersetup{
pdftitle={Python Agent in Ludii},
pdfsubject={Inter-process Communication with Python and Java},
pdfauthor={Izaias Saturnino de Lima Neto},
pdfkeywords={Inter-process communication},
}

\begin{document}
\maketitle

\begin{abstract}
Ludii is a Java general game system with a considerable number of board games, with an API for developing new agents and a game description language to create new games. To improve versatility and ease development, we provide Python interfaces for agent programming. This allows the use of Python modules to implement general game playing agents.

As a means of enabling Python for creating Ludii agents, the interfaces are implemented using different Java libraries: jpy and Py4J. The main goal of this work is to determine which version is faster. To do so, we conducted a performance analysis of two different GGP algorithms, Minimax adapted to GGP and MCTS. The analysis was performed across several combinatorial games with varying depth, branching factor, and ply time. For reproducibility, we provide tutorials and repositories.

Our analysis includes predictive models using regression, which suggest that jpy is faster than Py4J, however slower than a native Java Ludii agent, as expected.
\end{abstract}

% keywords can be removed
%\keywords{Inter-process communication}

\section{Introduction}

Ludii is a comprehensive game platform built to play, evaluate, and create various types of games (\cite{ludii}).
It is implemented in Java, offering an application programming interface (API) for agent development\footnote{Ludii API documentation: \url{https://ludiitutorials.readthedocs.io/en/latest/basic_ai_api.html}} and a game description language for game creation\footnote{Ludii language reference: \url{https://www.ludii.games/downloads/LudiiLanguageReference.pdf}}.

Interoperability between languages is beneficial in Ludii. For example, Python has numerous libraries for data analysis and machine learning. However, the overhead of the Python-Java communication can be considerable. Thus, the main goal of this work is to evaluate two different Python-Java communication methods: jpy\footnote{Jpy Git repository: \url{https://github.com/jpy-consortium/jpy}} and Py4J\footnote{Py4J Documentation: \url{https://www.py4j.org/contents.html}}.

The Python-Java bridges were evaluated by assessing the execution performance of a general game playing (GGP) implementation of the Minimax algorithm with alpha-beta pruning (\cite{alpha-beta-pruning}), along with Upper Confidence Bound for Trees (UCT) (\cite{uct}), a widely used variant of Monte Carlo Tree Search (MCTS). The selected games for this evaluation are combinatorial, defined as deterministic, two-player, zero-sum games with complete information. Different performance metrics were employed for each algorithm: the number of expanded states for Minimax and the number of playouts for UCT. The evaluation was conducted across 30 games with varying depth, ply time, and branching factor. The data collected from this analysis was used to develop predictive models. Consequently, the jpy agent was found to be faster than the Py4J agent but slower than the native Java agent, as anticipated. Tutorials and repositories for the jpy\footnote{Jpy agents implementations: \url{https://github.com/izaias-saturnino/jpy-ludii-example}} and Py4J\footnote{Py4J agents implementations: \url{https://github.com/izaias-saturnino/Py4j-ludii-example}} agents implementations are publicly available.

The report is structured as follows: Section \ref{sec:background} explains the concepts and tools used as basis for this report, briefly discusses how they function and how the Ludii agents are created. Section \ref{sec:performance} explains how the libraries are evaluated, presents the evaluation results, and discusses them. Section \ref{sec:prediction} shows the regression analysis, which provides insights into the behavior of the algorithms given a game's characteristics and can be used to predict the performance on new games. Finally, Section \ref{sec:conclusion} gives a brief overview, along with the findings and their significance.

\section{Background}
\label{sec:background}

General game playing artificial intelligence refers to a program capable of playing a wide variety of games. As explained by \cite{aaaicompetition} and further emphasized by \cite{hybriddiffadjust}, these algorithms cannot be tailored for specific games, since they receive a game description in some representation at runtime and must use this description to play the game effectively.
Several algorithms for general game playing AI exist, implementing different approaches to the problem. This report uses common and simple algorithms for the purposes of performance evaluation, which are further discussed in Sections \ref{sec:mcts} and \ref{sec:minimax}.

\subsection{Monte Carlo Tree Search}
\label{sec:mcts}

As the name suggests, Monte Carlo Tree Search (MCTS) is a tree search algorithm. There are several variants of this algorithm (\cite{mctssurvey}). 
MCTS is an iterative algorithm with four steps in each iteration: selection, expansion, simulation, and backpropagation.
The selection step chooses the leaf node on the partially-built game tree with the highest evaluation, where this evaluation is usually a function of the number of node visits and the reward associated with it. The next step is expansion, where a child of the selected leaf node is added to the tree. In the simulation step, a random game is played from the added node and backpropagation uses the result obtained on the simulation to update the number of visits and reward associated with each node traversed in the tree. The intuition is that the selection process directs the search towards more promising tree regions, building an unbalanced tree tending towards stronger moves. Since the algorithm uses statistics, it may return a suboptimal move with finite time. Even so, it tends to choose better moves given more time.

This report uses the Upper Confidence bounds for Trees (UCT) variant of MCTS (\cite{uct}), which is simple to implement and understand. During selection, UCT favors nodes with the highest average reward plus its upper confidence bound, related to how frequently the node was visited. Given infinite time, this variant is guaranteed to infer the path that maximizes the reward.

\subsection{Minimax}
\label{sec:minimax}

Minimax is another tree search algorithm. It differs significantly from UCT, since it calculates the value of the current state without the use of probability, requiring the construction of the entire subtree from the current state in a depth-first fashion. This allows the discovery of all terminal node values and the propagation of this information upwards. Since the tree grows exponentially, finding the true state value is unfeasible in most cases.

A variation of this algorithm is the alpha–beta pruning, which reduces the number of evaluated states without losing optimality (\cite{alpha-beta-pruning}). The Minimax algorithm variation used is an anytime variation with alpha-beta pruning and no heuristic. When the time limit is reached, the algorithm assumes that every non-evaluated state is a draw for the purpose of selecting a move.

\subsection{Ludii}
\label{sec:ludii}

Ludii is a general game system designed to play, evaluate, and create a wide range of games (\cite{ludii}). Ludii provides a game description language for creating games and an API for developing agents.

A game description language serves as a formalism for specifying the rules, objects, and behavior of a game in a structured format. By using the game description language, a game can be expressed in a concise and formal manner, enabling its realization and evaluation within the Ludii system. This language abstracts away the complexities of a programming language, allowing a focus on the conceptual aspects of game design.

Ludii's API for developing agents provides a set of functions and protocols that enable the creation of intelligent agents capable of playing the games within Ludii. This API acts as a bridge between the game engine and third-party code, supporting agents with custom strategies and algorithms.

\subsection{Python-Java inter-process communication}
\label{sec:communication}

When creating Python agents for Ludii, communication is necessary between the Java Virtual Machine (JVM) running Ludii and the Python interpreter running the agent. This communication is established using two different Java libraries, jpy and Py4J. In this work, we evaluate Ludii agents implementing anytime Minimax alpha-beta pruning and UCT in Python on both Python-Java bridges. The UCT agents are based on the Ludeme project agents\footnote{Ludeme Python agent repository: \url{https://github.com/Ludeme/LudiiPythonAI}}\footnote{Ludeme Java agent repository: \url{https://github.com/Ludeme/LudiiExampleAI}}. The Minimax agents are based on the original alpha-beta pruning variant, with modifications to exclude heuristics and enable them to function as anytime agents.

When implementing an agent with jpy, the jpy library source code was cloned from jpy's Git repository and built. Afterwards, the dependencies for the jpy project were added manually. To demonstrate how to create a working environment, we created a Git repository\footnote{Jpy agent Git repository: \url{https://github.com/izaias-saturnino/jpy-ludii-example}} with the locally working project and a step-by-step guide. Simply cloning the repository may not yield the expected result, as the project was built for a specific environment. In other words, the steps must be followed to ensure the project functions properly in different configurations. The relevant characteristics of the experimental environment include Windows 11, Python 3.10, and Java 21.0.1, along with 16GB of RAM and an AMD Ryzen X Series processor featuring a clock speed of 3.6 GHz, 6 cores, and 12 threads.

For the Py4J agent, the Py4J Python module was installed using pip, and the Java library was built automatically. A Git repository with the Py4J project and a step-by-step guide is also provided\footnote{Py4J agent Git repository: \url{https://github.com/izaias-saturnino/Py4j-ludii-example}}.

Py4J uses a server to exchange data. Consequently, if the JVM starts the server, as in this implementation, the separate Python interpreter process that implements the agent must be executed before the agent is used by Ludii. Another consequence is the overhead caused by the network layer in communication between the JVM and the Python interpreter, which could be avoided by using other communication methods.

\section{Performance Analysis}
\label{sec:performance}

To evaluate the different libraries, we implemented each GGP algorithm using both libraries, with the only difference between the implementations being the Java-Python communication (i.e., the library itself). We also created an equivalent native Java agent to further improve the experiment and compare performance against a Java-only implementation. The GGP algorithms used were UCT (Section \ref{sec:mcts}) and an anytime Minimax with alpha-beta pruning and no heuristic (Section \ref{sec:minimax}). In a preliminary analysis, the agents were evaluated in Chess, Go, and Shogi. The respective source codes for the implementations are available in the Git repositories previously mentioned. For the Java implementation, refer to ExampleUCT.java and Minimax.java, and for the Python implementations, refer to uct.py and minimax.py in both repositories.

Different evaluation metrics were used for each algorithm. For the UCT agents, we evaluated the number of playouts in a fixed time. For the Minimax agents, we evaluated the number of expanded game states in 1 second. These metrics are intrinsically related to the nature of these algorithms: for UCT, more playouts provide more accurate estimates of action values, whereas for Minimax, more accurate estimates are obtained as the game tree is explored more deeply.
We also conducted an average score evaluation for each agent. This average score was calculated with a win as 1, a draw as 0.5, and a loss as 0.

\begin{figure}[htbp]
	\centering
	\includegraphics[width=0.5\textwidth]{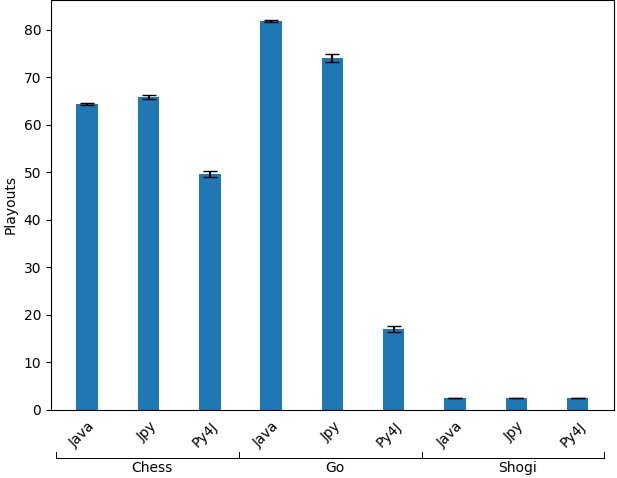}
	\caption{Average number of playouts performed per second in MCTS. The 99\% confidence interval (CI) is represented by the error bar.}
	\label{fig:playouts_analysis}
\end{figure}

The comparison of the number of playouts used the average and standard deviation of playouts on the first move of a game when the agent was the first player. For each agent, 1000 evaluations were made. In Chess and Go, all agents had 1 second to make their move. For Shogi, only 2 playouts were performed in 1 second. Hence, the time used for evaluation in Shogi was 10 seconds. This difference in Shogi evaluation time arises from the relatively small number of playouts performed in 1 second, which would affect the standard deviation and the analysis results. The resulting average number of playouts in Shogi was subsequently divided by 10. The playout analysis can be found in Figure \ref{fig:playouts_analysis}.

The low variability in the performance of all implemented UCT agents in Shogi may be attributed to the limited frequency of calls from the Python interpreter to the JVM. This occurs specifically in Shogi because the game has a larger average depth for a random player.

\begin{figure}[htbp]
	\centering
	\includegraphics[width=0.5\textwidth]{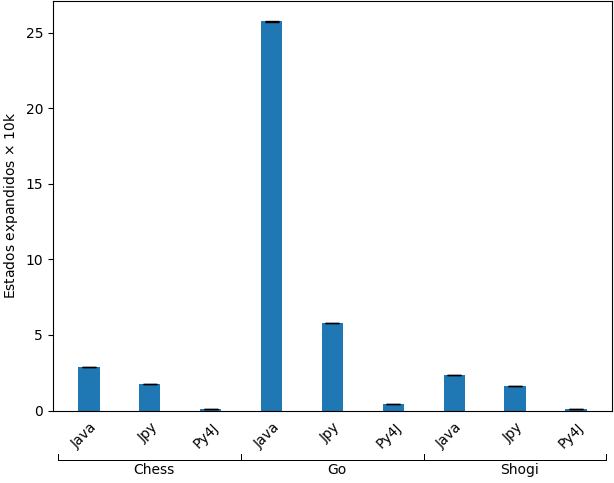}
	\caption{Average number of expanded states in Minimax per second. The 99\% confidence interval (CI) is represented by the error bar.}
	\label{fig:expanded_states_analysis}
\end{figure}

The state expansion comparison used the same methodology as the playout comparison. For the first move of every game, the average and confidence interval (CI) of the number of expanded states were measured. For each agent, 1000 evaluations were made. In Chess and Go, all agents had 1 second to play their move. For Shogi, to match the time used in MCTS, 10 seconds were allocated. The resulting average number of expanded states in Shogi was also divided by 10 due to the difference in allotted time.
The state expansion analysis can be found in Figure \ref{fig:expanded_states_analysis}.

In the score comparison, 30 Chess games were played, alternating the starting player. The agents used were only UCT, as Minimax without heuristics would not be able to evaluate initial and mid-game states. All agents had one second to make their moves. The average score was calculated, awarding 1 point for a win, 0.5 for a draw, and 0 for a loss. The score analysis can be found in Table \ref{tab:score}.

\begin{table}[H]
	\caption{Score analysis. A total of 30 games were played for each comparison. Since the UCT jpy agent defeated its Py4J counterpart, it was then compared to the native Java implementation, which proved superior.}
	\centering
	\begin{tabular}{lllll}
		\toprule
		Player 1 & Player 2 & Highest scoring player & Score average \\
		\midrule
		UCT jpy  & UCT Py4J & UCT jpy  & 0.65 \\
		UCT jpy  & UCT Java & UCT Java & 0.5333 \\
		\bottomrule
	\end{tabular}
	\label{tab:score}
\end{table}

The results show a higher average score for the jpy player against the Py4J player, while the Java player recorded a higher average score than the Jpy player, by a smaller margin. These findings align with the analysis of the number of playouts.

Since 3 games were used for the number of playouts analysis and only one for win rate evaluation, the conclusions drawn from these may not generalize to other games. Thus, in a secondary analysis, 100 more games were analyzed. These games selected were the initial entries of games in the list of games in Ludii. These games are combinatorial, classified as deterministic, two-player, zero-sum games with complete information. All of the selected games belong to the collections of hunt and escape games. From these, statistics on their game complexity were obtained through Ludii itself. Based on this information, a subset of 30 games was selected to analyze the impact of varying the obtained metrics. This subset was most suitable for the purpose, as it consists of groups of games that vary primarily in one of the game complexity metrics, allowing for a correct prediction of variations in individual metrics.

For each game, its performance on each of the implementations was measured. This was done in the same manner as explained in the performance analysis section, except that the number of evaluations from the starting state in 1 second was 100 instead of 1000.

\begin{figure}[htbp]
    \centering
    \subcaptionbox{Java heatmap of rollouts \label{fig:heatmap_java_mcts}}{
        \includegraphics[width=0.48\textwidth]{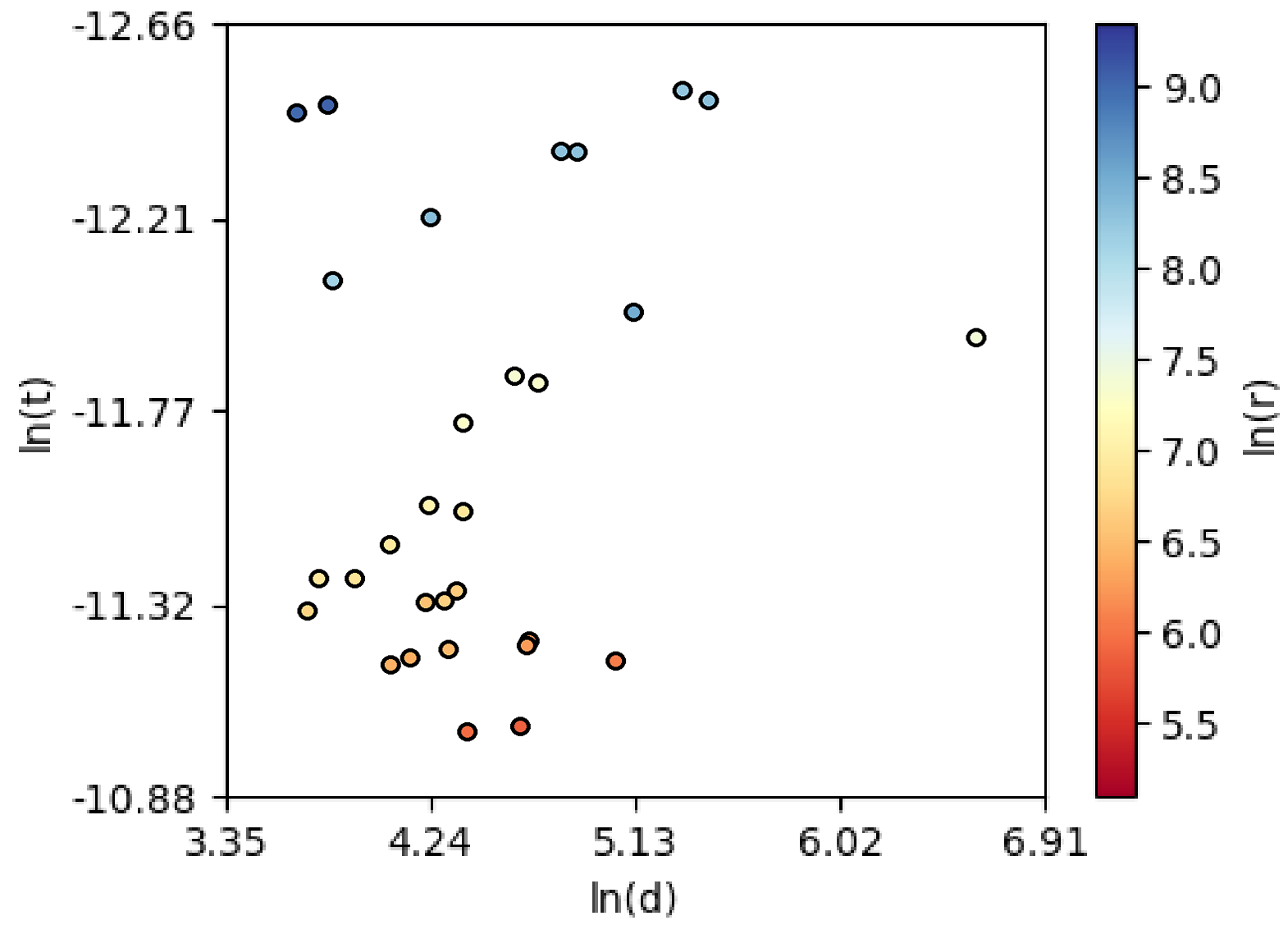}
    }
    \subcaptionbox{Java heatmap of expanded states \label{fig:heatmap_java_minimax}}{
        \includegraphics[width=0.48\textwidth]{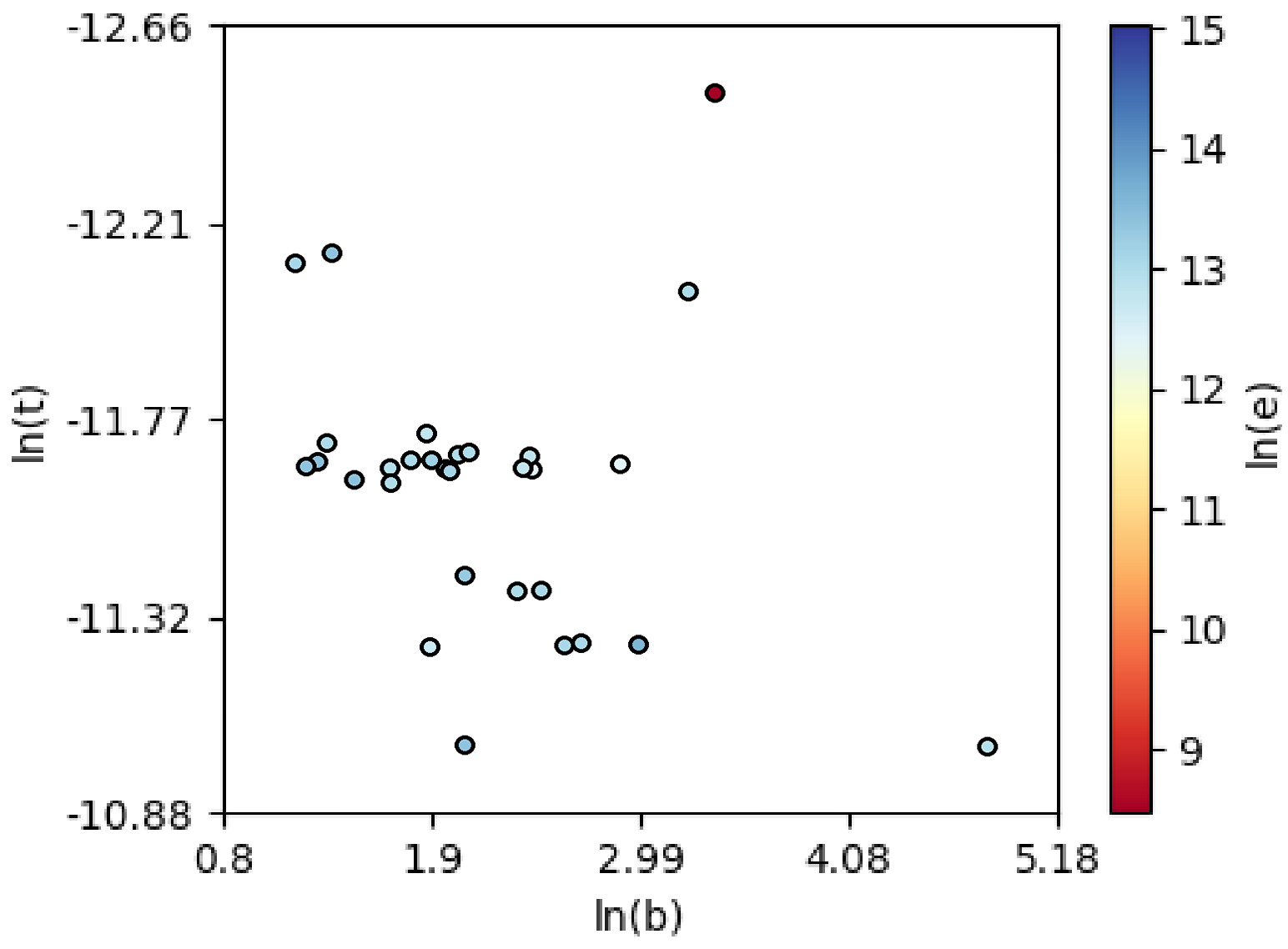}
    }
    \caption{Java data heatmaps}
\end{figure}

\begin{figure}[htbp]
    \centering
    \subcaptionbox{Jpy heatmap of rollouts \label{fig:heatmap_jpy_mcts}}{
        \includegraphics[width=0.48\textwidth]{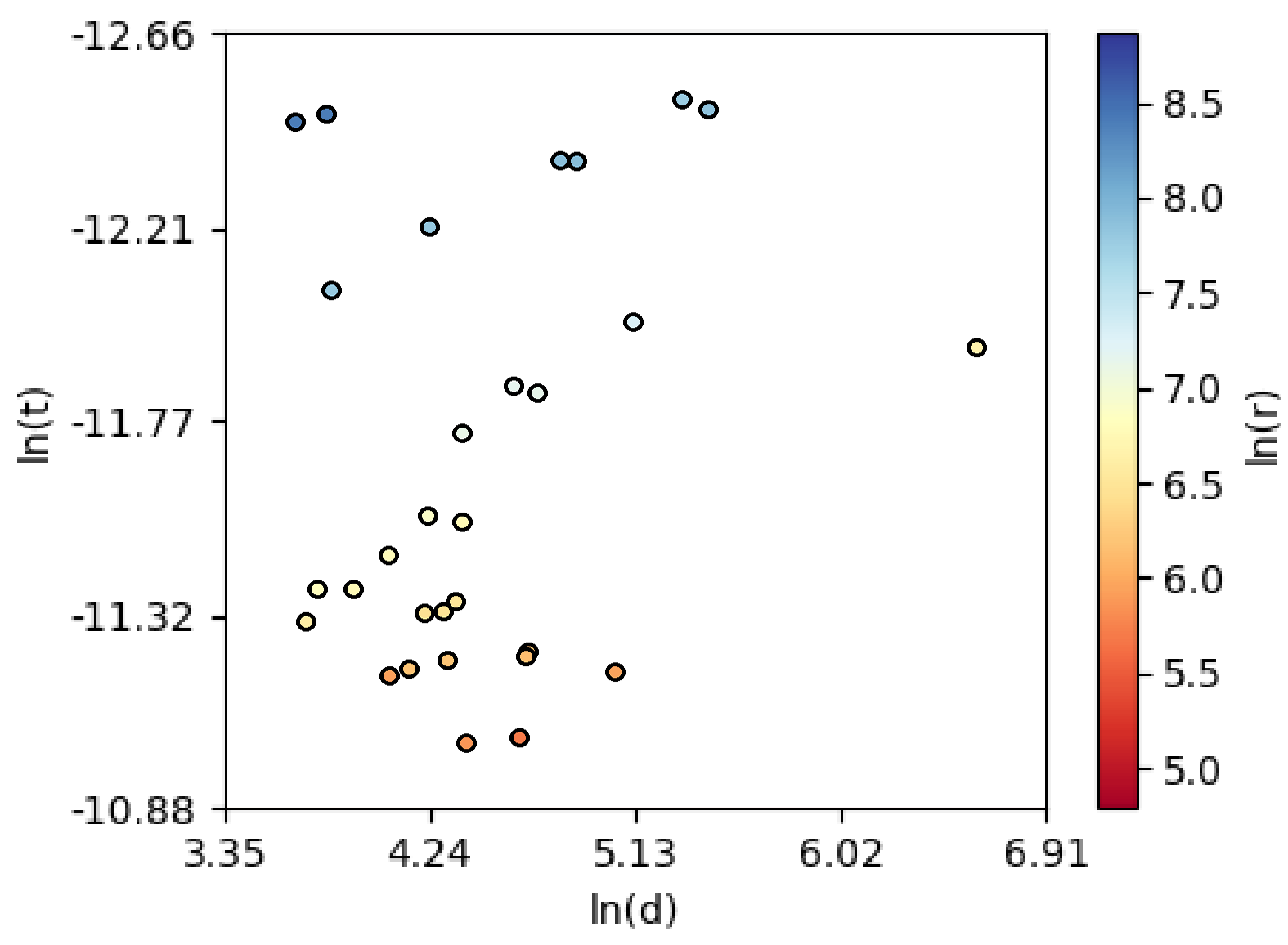}
    }
    \subcaptionbox{Jpy heatmap of expanded states \label{fig:heatmap_jpy_minimax}}{
        \includegraphics[width=0.48\textwidth]{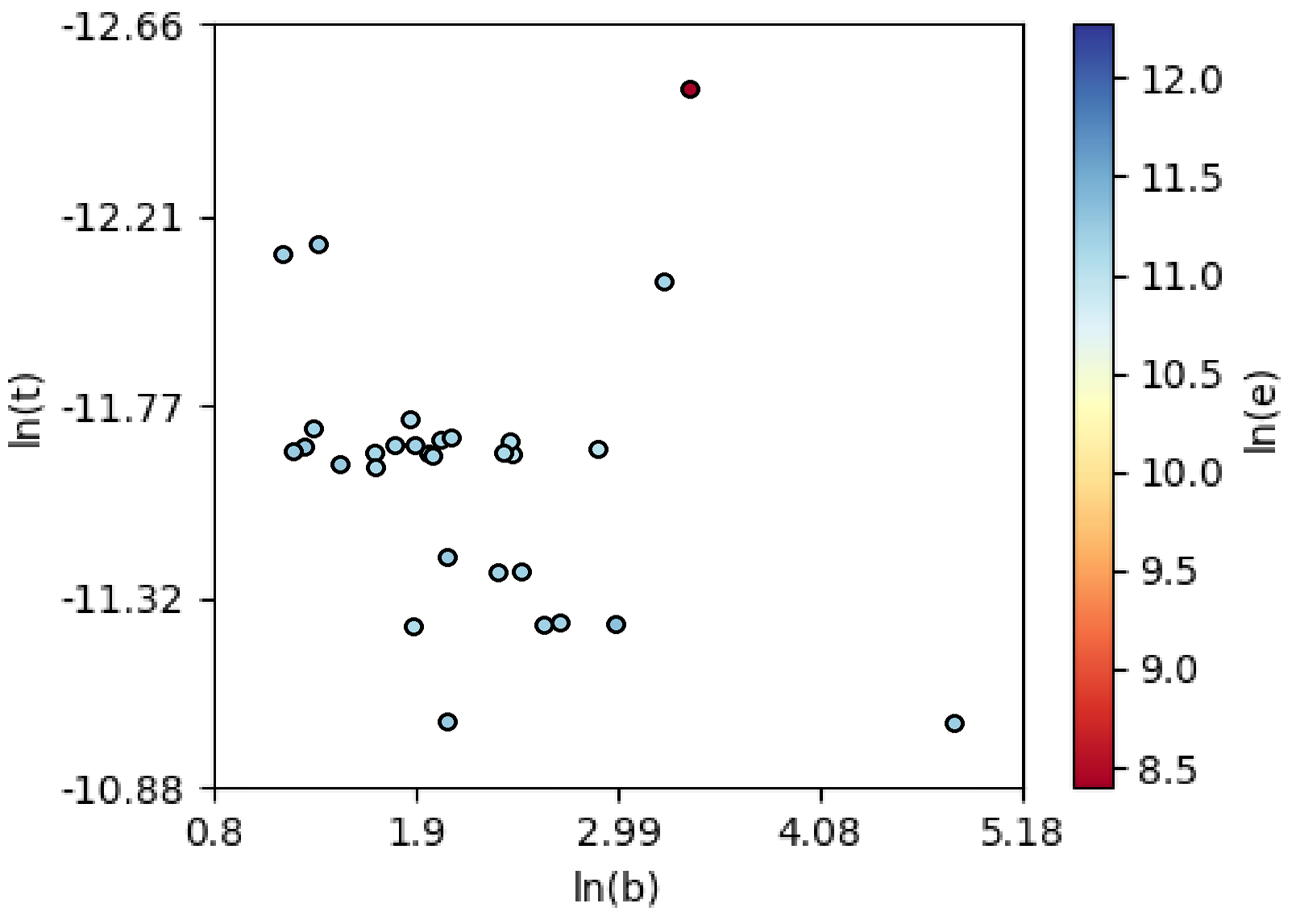}
    }
    \caption{Jpy data heatmaps}
\end{figure}

\begin{figure}[htbp]
    \centering
    \subcaptionbox{Py4J heatmap of rollouts \label{fig:heatmap_py4j_mcts}}{
        \includegraphics[width=0.48\textwidth]{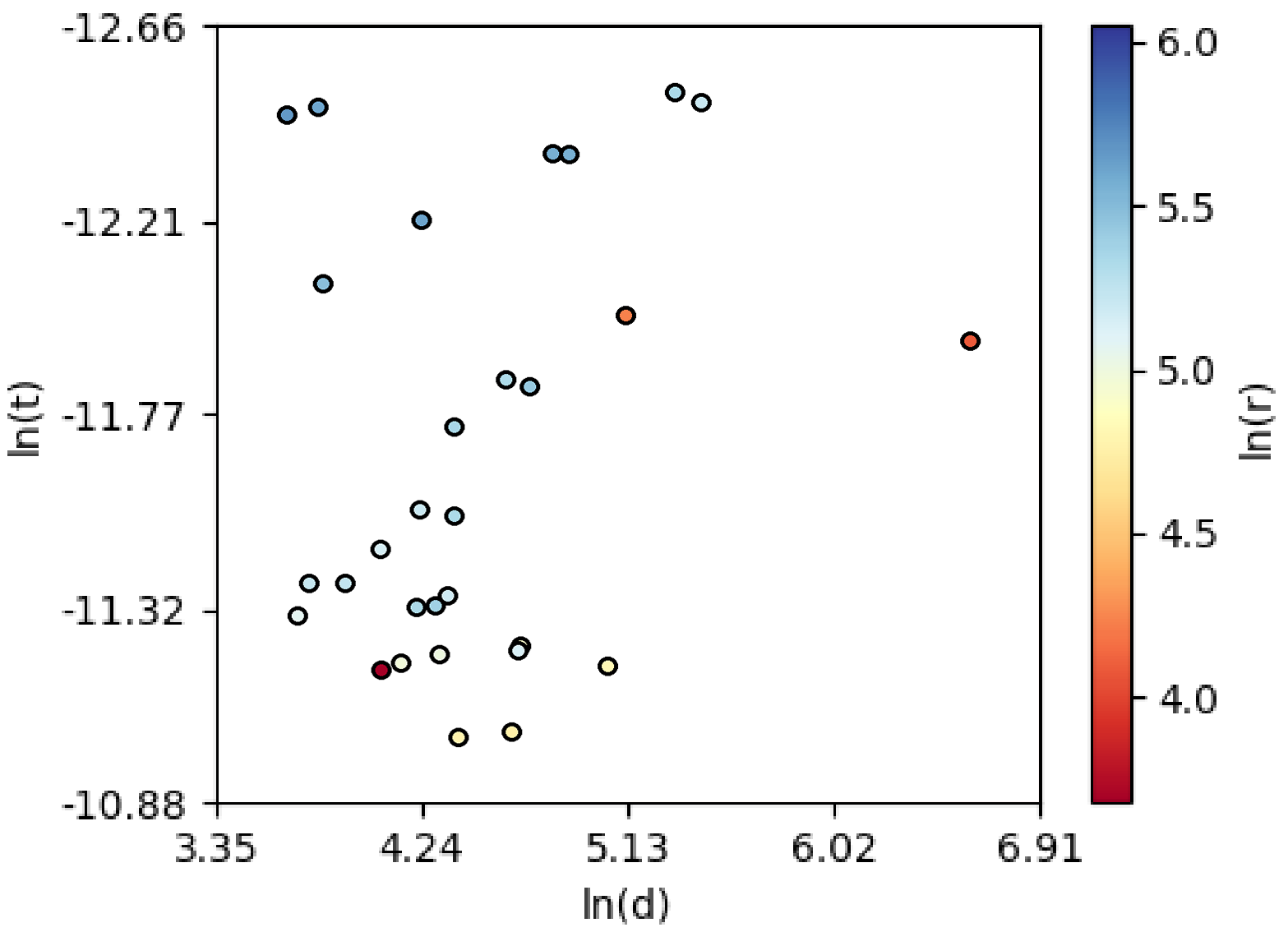}
    }
    \subcaptionbox{Py4J heatmap of expanded states \label{fig:heatmap_py4j_minimax}}{
        \includegraphics[width=0.48\textwidth]{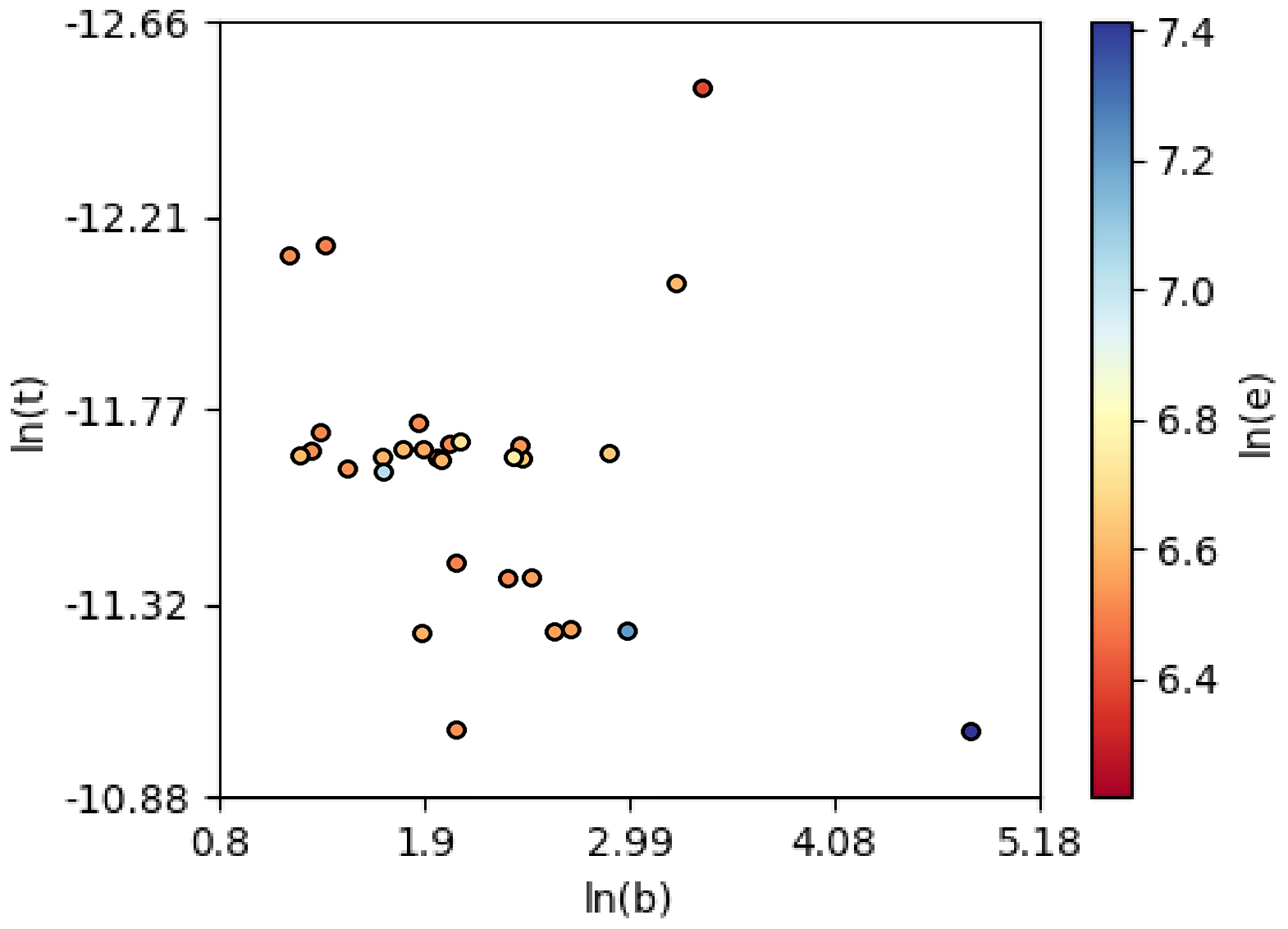}
    }
    \caption{Py4J data heatmaps}
\end{figure}

Figures \ref{fig:heatmap_java_mcts}, \ref{fig:heatmap_java_minimax}, \ref{fig:heatmap_jpy_mcts}, \ref{fig:heatmap_jpy_minimax}, \ref{fig:heatmap_py4j_mcts}, and \ref{fig:heatmap_py4j_minimax} present heatmaps that illustrate the performance of each game in relation to the number of MCTS rollouts (r), expanded states of Minimax (e), game depth (d), ply time (t), and branching factor (b).

As observed in the data, the number of playouts tends to be negatively impacted by the game depth and ply time in MCTS. In Minimax, the number of expanded states tends to be negatively impacted by the ply time and positively impacted by the branching factor, which is an expected effect of using the number of expanded states as a performance metric. This is not entirely reflected in the Py4J implementation of Minimax, as the branching factor negatively impacts its performance. This is believed to occur due to Py4J's considerably lower performance with this algorithm.

\section{Performance Prediction}
\label{sec:prediction}

This section presents a regression analysis aimed at evaluating the influence of game features on the performance of MCTS and Minimax algorithms. Specifically, the following features are considered: game depth (d), which refers to the average number of moves required to complete a game; ply time (t), defined as the average time taken to process a move; and branching factor (b), representing the average number of possible moves at each game state. The objective of this analysis is to predict the performance of MCTS and Minimax in terms of the number of playouts (or rollouts, r) and the number of expanded states (e), respectively, for their implementations using native Java, jpy, and Py4J.

For the purpose of this analysis, the ply time employed corresponds to the Java ply time. Given that ply time exhibits exponential behavior in the regression model, the specific ply time utilized does not significantly influence the results. This is due to the fact that taking the logarithm of a value in one base and subsequently converting to another base yields the same outcome, but scaled by a constant factor. While using varying ply times is a valid approach for analyzing behavioral patterns, it is not deemed critical for accurately predicting performance.

The regressions were conducted with a logarithmic transformation applied to all terms. Regressions using linear terms were also tested; however, these yielded less significant results, indicating an underlying exponential behavior. This is consistent with expectations, as the growth of the game tree is exponential, which makes logarithmic transformations particularly appropriate.

These equations were derived using gradient descent. More information regarding these regressions, such as their mean squared error (MSE), is available in Table \ref{tab:equations}.

\begin{table}[H]
	\caption{Regressions with logarithmic transformation conducted to predict the number of rollouts in MCTS (r) and the number of expanded states in Minimax (e) based on a game's depth (d), ply time (t), and branching factor (b), alongside their MSE and MSE in the test set. Some terms were excluded from the regressions because their coefficients were found to be close to zero in a model that included all terms. As a result, these terms were removed from the final regression model.}
	\centering
	\begin{tabular}{lllll}
		\toprule
		Algorithm & Implementation & Equation & MSE & MSE test \\
		\midrule
		MCTS & Java & ln(r) = -0.946*ln(d) - 0.492*ln(t) + 6.291 & 0.0357 & 0.028 \\
        %MCTS & Java & ln(r) = -0.96*ln(d) + 0.071*ln(b) - 0.546*ln(t) + 5.472 & 0.03314 & \\
		MCTS & Jpy & ln(r) = -0.801*ln(d) - 0.686*ln(t) + 2.87 & 0.00862 & 0.007 \\
        %MCTS & Jpy & ln(r) = -0.809*ln(d) + 0.045*ln(b) - 0.72*ln(t) + 2.353 & 0.007566 & \\
        MCTS & Py4J & ln(r) = -0.271*ln(d) - 0.787*ln(t) - 3.005 & 0.1243 & 0.054 \\ \midrule
        %MCTS & Py4J & ln(r) = -0.27*ln(d) - 0.005*ln(b) - 0.784*ln(t) - 2.951 & 0.1243 & \\
        Minimax & Java & ln(e) = 0.818*ln(b) - 1.665*ln(t) - 10.291 & 0.45 & 1.738 \\
        %Minimax & Java & ln(e) = -0.035*ln(d) + 0.829*ln(b) - 1.7*ln(t) - 10.551 & 0.449036 & - \\
		Minimax & Jpy & ln(e) = 0.513*ln(b) - 0.914*ln(t) - 1.95 & 0.18084 & 0.515 \\
        %Minimax & Jpy & ln(e) = 0.027*ln(d) + 0.504*ln(b) - 0.888*ln(t) - 1.75 & 0.18022 & - \\
        Minimax & Py4J & ln(e) = 0.11*ln(b) + 0.399*ln(t) + 11.182 & 0.043953 & 0.099 \\
        %Minimax & Py4J & ln(e) = 0.045*ln(d) + 0.096*ln(b) + 0.443*ln(t) + 11.516 & 0.04222 & - \\
		\bottomrule
	\end{tabular}
	\label{tab:equations}
\end{table}

For the test set, three additional games were selected, Asalto, El Zorro and Janes Soppi. This test allows for the verification that the regressions provide accurate approximations of performance behavior. By utilizing the equations from Table \ref{tab:equations}, reliable approximations of the true values can be obtained for hunt and escape games with comparable game complexities, i.e., complexities similar to those used in the creation of the models. This does not apply to games such as Chess, Shogi, and Go, as their game complexities differ significantly from those of the other games analyzed.

\begin{figure}[htbp]
    \centering
    \subcaptionbox{Java heatmap of rollouts regression \label{fig:reg_heatmap_java_mcts}}{
        \includegraphics[width=0.48\textwidth]{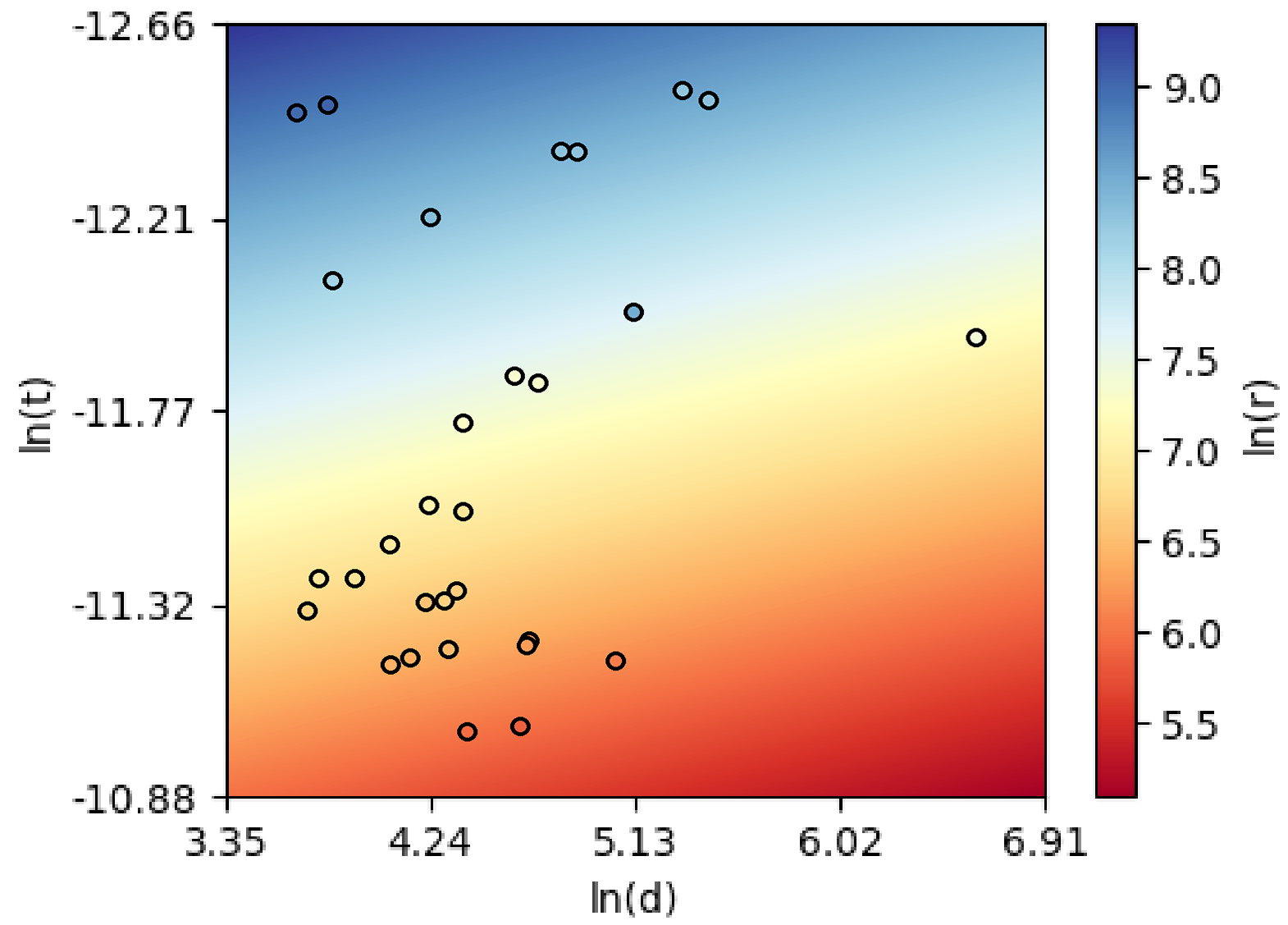}
    }
    \subcaptionbox{Java heatmap of expanded states regression \label{fig:reg_heatmap_java_minimax}}{
        \includegraphics[width=0.48\textwidth]{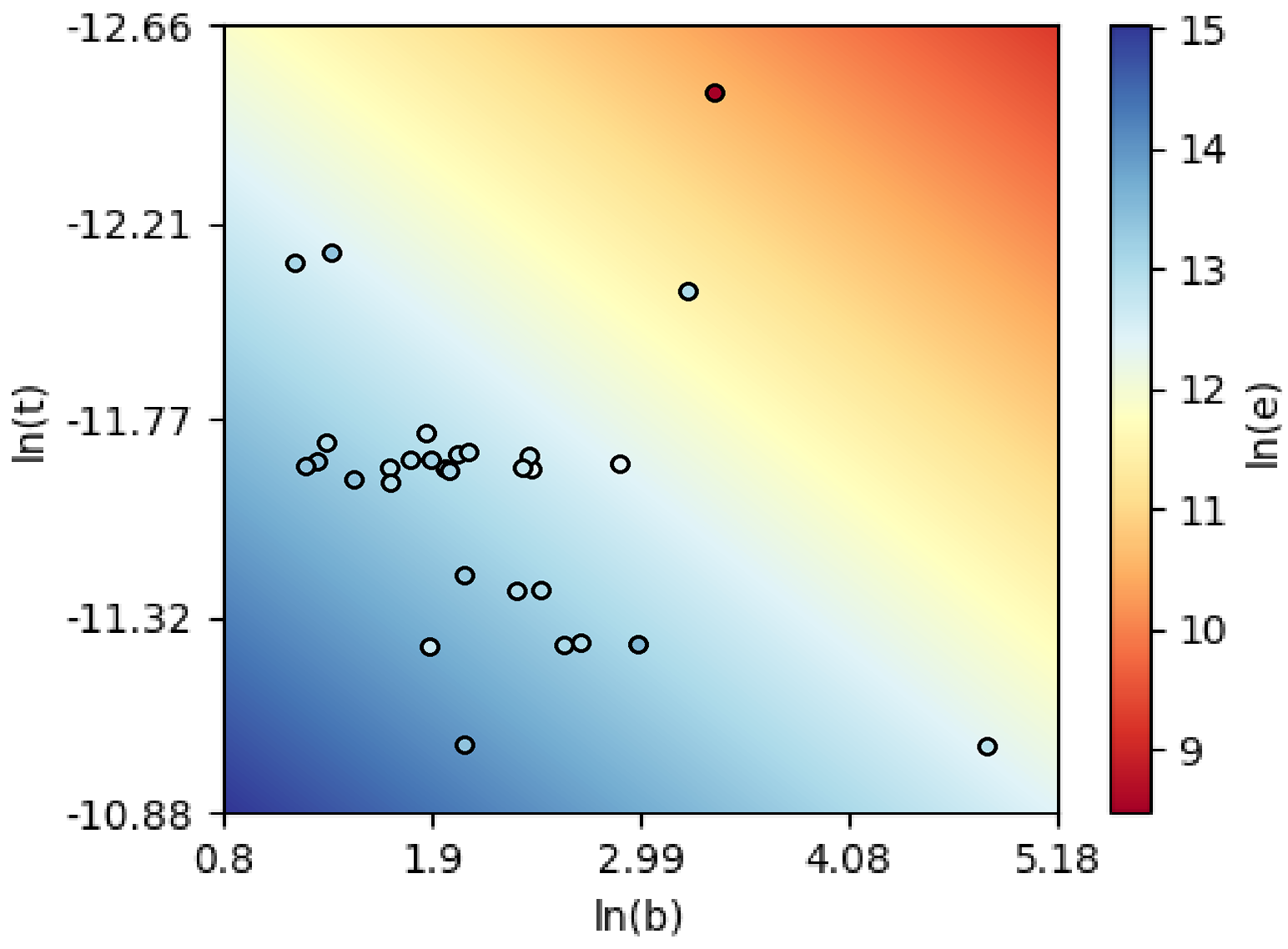}
    }
    \caption{Java regressions heatmaps}
\end{figure}

\begin{figure}[htbp]
    \centering
    \subcaptionbox{Jpy heatmap of rollouts regression \label{fig:reg_heatmap_jpy_mcts}}{
        \includegraphics[width=0.48\textwidth]{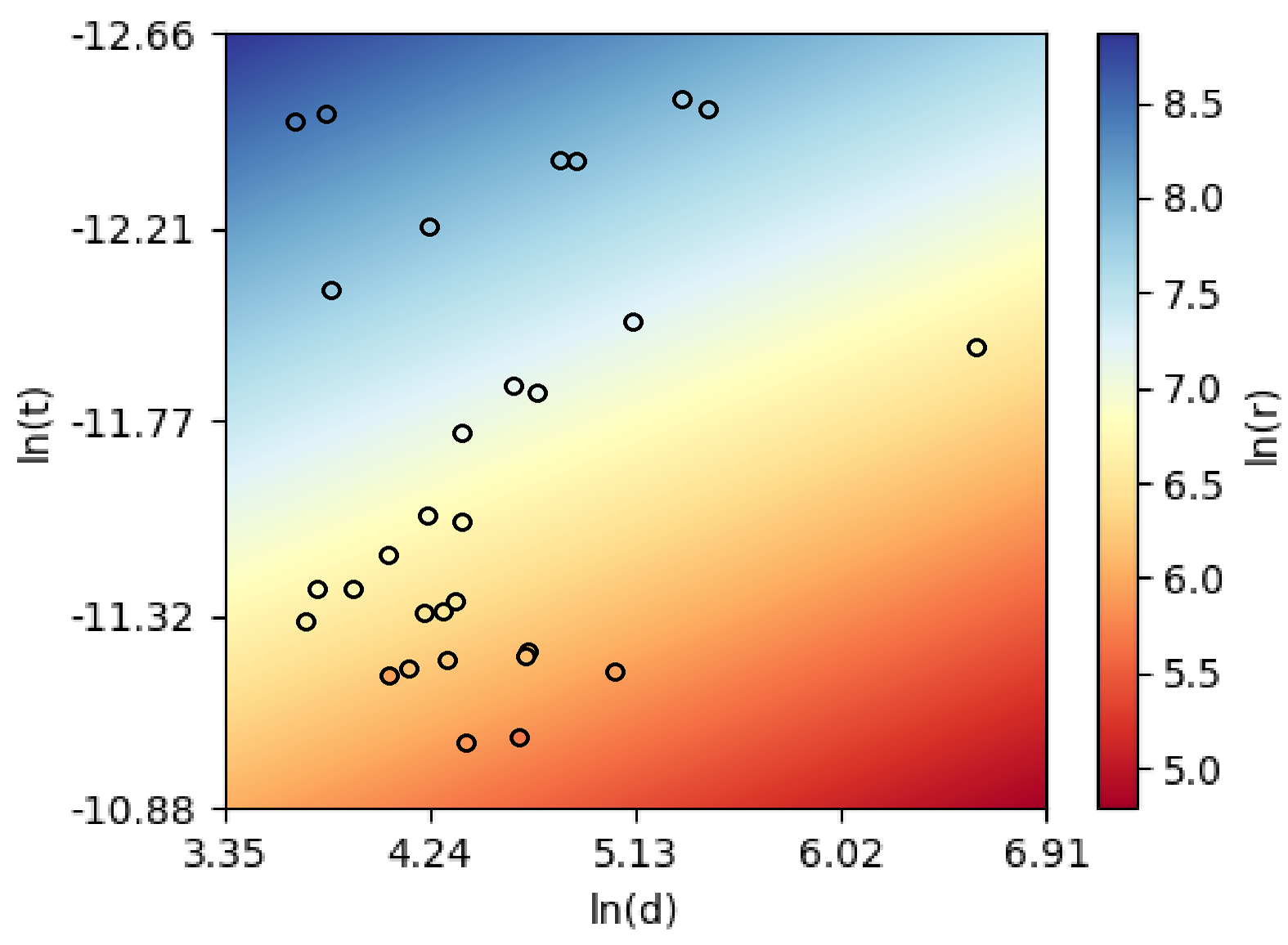}
    }
    \subcaptionbox{Jpy heatmap of expanded states regression\label{fig:reg_heatmap_jpy_minimax}}{
        \includegraphics[width=0.48\textwidth]{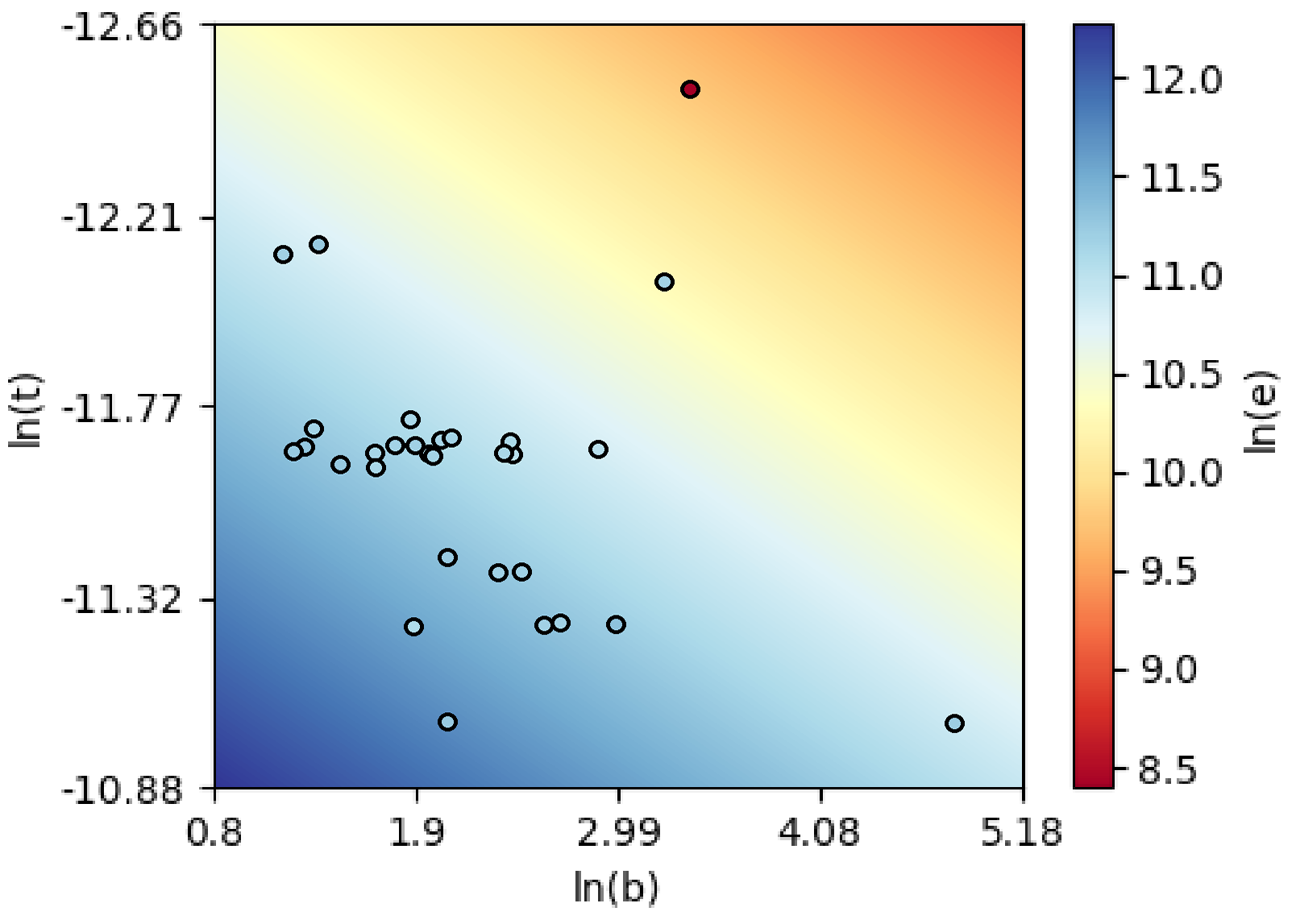}
    }
    \caption{Jpy regressions heatmaps}
\end{figure}

\begin{figure}[htbp]
    \centering
    \subcaptionbox{Py4J heatmap of rollouts regression \label{fig:reg_heatmap_py4j_mcts}}{
        \includegraphics[width=0.48\textwidth]{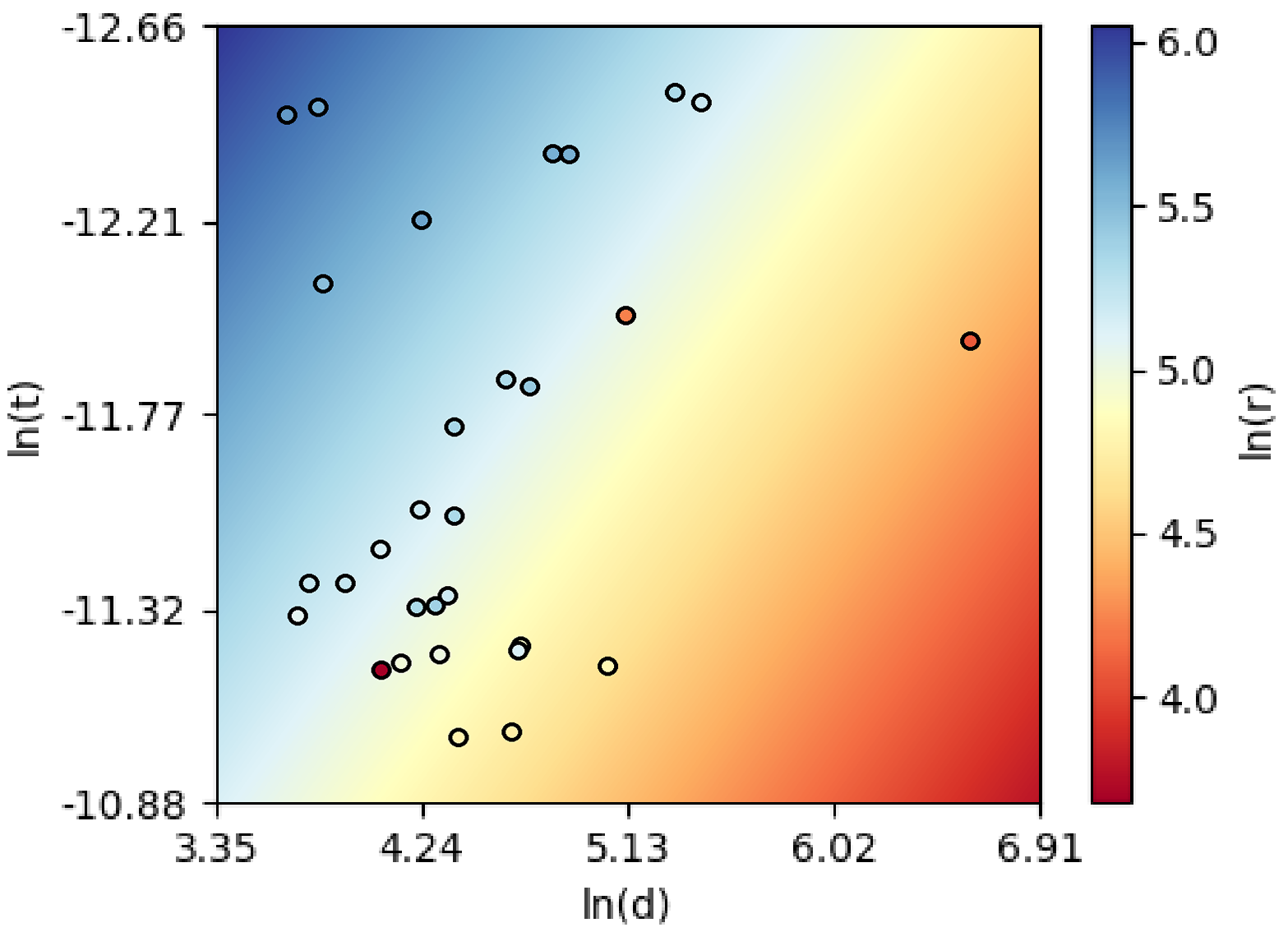}
    }
    \subcaptionbox{Py4J heatmap of expanded states regression \label{fig:reg_heatmap_py4j_minimax}}{
        \includegraphics[width=0.48\textwidth]{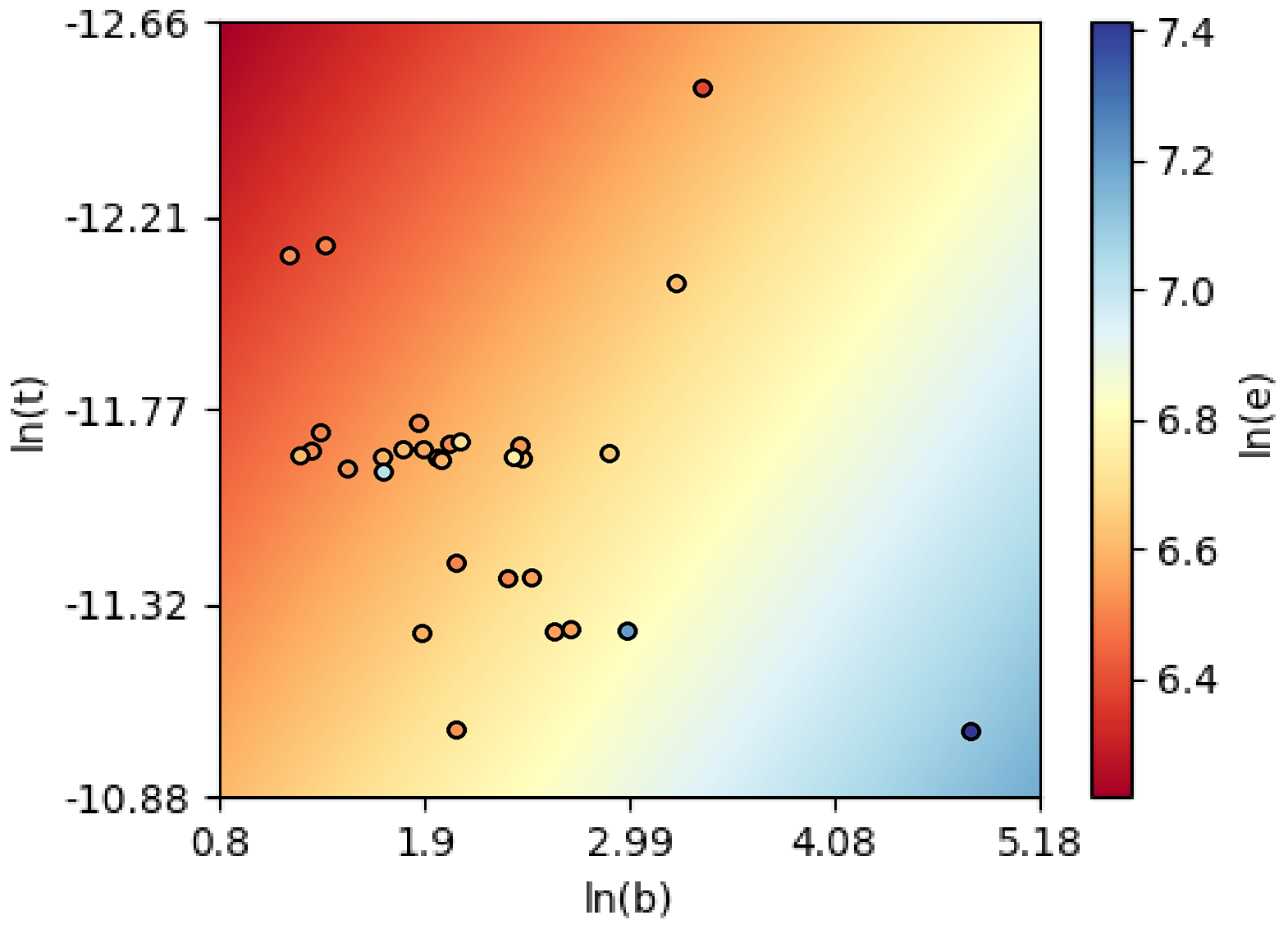}
    }
    \caption{Jpy regressions heatmaps}
\end{figure}

Figures \ref{fig:reg_heatmap_java_mcts}, \ref{fig:reg_heatmap_java_minimax}, \ref{fig:reg_heatmap_jpy_mcts}, \ref{fig:reg_heatmap_jpy_minimax}, \ref{fig:reg_heatmap_py4j_mcts}, and \ref{fig:reg_heatmap_py4j_minimax} present heatmaps that depict performance regressions in relation to various factors: the number of MCTS rollouts (r), the expanded states of Minimax (e), game depth (d), ply time (t), and branching factor (b). These heatmaps also display the data points, and errors can be easily recognized by observing the color of the surface. A noticeable variation between the color of a data point and the surrounding surface color suggests the presence of a significant error.

The obtained regressions are effective in predicting a game's behavior based on its complexity, implementation method, and algorithm. This is evident from the relatively low MSE on the test set. Additionally, the regressions help explain why certain games exhibit superior performance with specific algorithms or implementation methods. Since the impact of variables is correlated with their coefficients, the equations suggest that Java implementations outperform jpy implementations, which, in turn, surpass Py4J implementations.

Regressions that incorporated depth, ply time, and branching factor simultaneously were conducted for each algorithm and implementation. However, no significant difference in MSE was observed. The coefficients for some terms were found to be close to zero and, as a result, were excluded from the final regression model. This outcome may be attributed to the minimal or nonexistent relevance of these additional variables.

The regressions may not accurately capture points that deviate from the majority, possibly due to measurement error or limitations in generalizing for certain types of games. Nevertheless, they remain valuable for understanding the overall patterns of similar data points.

\section{Conclusion}
\label{sec:conclusion}

This technical report provides a comparative analysis of two Java-Python communication libraries, jpy and Py4J, within the context of Python-based GGP agents for Ludii, a general game system developed in Java. The UCT and Minimax algorithms were implemented using each communication library to evaluate the extent to which the choice of library influences the performance of the algorithms. Furthermore, predictive models were constructed to estimate the performance of these implementations based on game characteristics.

We conclude that performance improvements are closely correlated with the frequency of inter-process communication events, as communication between processes varies with the algorithm's design. UCT communicates between processes each time it executes a rollout and performs multiple rollouts, whereas Minimax does so a fixed number of times per node, which results in UCT being significantly impacted by communication overhead. Furthermore, Java implementations exhibit superior performance relative to jpy implementations, which in turn outperform Py4J implementations.

A potential area for future exploration involves researching methods for communicating the game state to Python in a way that minimizes inter-process communication. This approach would enable the creation of Ludii agents entirely within Python, using jpy.

\section{Acknowledgements}
\label{sec:acknowledgements}

The first author would like to thank Instituto TIM and OBMEP for their financial support.

\nocite{jpy,py4j,ludeme}
\bibliographystyle{unsrtnat}
\bibliography{references}

\end{document}